\documentclass[times, review, 10pt]{elsarticle}

\usepackage[table]{xcolor}

\usepackage{amssymb}
\usepackage{amsmath}
\usepackage{amsfonts}       
\usepackage{nicefrac}       
\usepackage{microtype}      
\usepackage{amsthm}

\usepackage{booktabs}
\usepackage{multirow}
\usepackage{soul}
\usepackage{bm}
\usepackage{tikz}
\definecolor{deepred}{RGB}{192,0,0}
\definecolor{deepyellow}{RGB}{197,90,17}
\definecolor{deepblue}{RGB}{0,128,172}
\definecolor{deepgreen}{RGB}{84,130,53}
\definecolor{skyblue}{RGB}{0,191,255}
\definecolor{indigo}{RGB}{102,0,153}

\usepackage{graphicx}
\definecolor{lightyellow}{rgb}{1.00, 0.98, 0.94}
\definecolor{mix1}{rgb}{0.9665, 0.984, 0.964}
\definecolor{ourscolor}{HTML}{f3f9f1}
\definecolor{lightorange}{HTML}{f8cbad}
\definecolor{lightblue}{HTML}{8faadc}
\definecolor{pink}{HTML}{ff66ff}
\definecolor{yellow}{HTML}{ffd966}

\usepackage{algorithmic}

\usepackage{algorithm}
\usepackage{caption}
\newcommand{\MyComment}[1]{\hfill \makebox[0.45\linewidth][l]{\textit{\color{gray}/* #1 */}}}

\newcommand{\revision}[1]{#1} 

\usepackage{hyperref}
\hypersetup{
    colorlinks=true,
    linkcolor=deepblue,
    citecolor=deepblue,
    filecolor=deepblue,
    urlcolor=deepblue,
}

\usepackage{makecell}

\newif\ifdraft 

\begin{document}

\begin{frontmatter}



\title{Tracking by Detection and Query: An Efficient End-to-End Framework for Multi-Object Tracking}


\author{Shukun Jia\textsuperscript{a,b}, Shiyu Hu\textsuperscript{c}, Yichao Cao\textsuperscript{d}, Feng Yang\textsuperscript{e}, Xin Lu\textsuperscript{a,b}, Xiaobo Lu\textsuperscript{a,b,*}}

\affiliation{organization={School of Automation},
            addressline={Southeast University}, 
            city={Nanjing},
            postcode={210096}, 
            country={China}}
\affiliation{organization={Key Laboratory of Measurement and Control of Complex Systems of Engineering},
            addressline={Ministry of Education}, 
            city={Nanjing},
            postcode={210096}, 
            country={China}}
\affiliation{organization={School of Physical \& Mathematical Sciences},
            addressline={Nanyang Technological University}, 
            postcode={639798}, 
            country={Singapore}}
\affiliation{organization={Big Data Institute},
            addressline={Central South University}, 
            city={Changsha},
            postcode={410083}, 
            country={China}}
\affiliation{organization={School of Instrument Science and Engineering},
            addressline={Southeast University}, 
            city={Nanjing},
            postcode={210096}, 
            country={China}}

\begin{abstract}
Multi-object tracking (MOT) is primarily dominated by two paradigms: tracking-by-detection (TBD) and tracking-by-query (TBQ). While TBD offers modular efficiency, its fragmented association pipeline often limits robustness in complex scenarios. Conversely, TBQ enhances semantic modeling end-to-end but suffers from high training costs and slow inference due to the tight coupling of detection and association. 
In this work, we propose the tracking-by-detection-and-query framework, TBDQ-Net, to advance the synergy between TBD and TBQ paradigms. By integrating a frozen detector with a lightweight associator, this architecture ensures intrinsic efficiency. Within this streamlined framework, we introduce tailored designs to address MOT-specific challenges.
Concretely, we alleviate task conflicts and occlusions through the dual-stream update of the Basic Information Interaction (BII) module. The Content-Position Alignment (CPA) module further refines both content and positional components, providing well-aligned representations for association decoding. Extensive evaluations on DanceTrack, SportsMOT, and MOT20 benchmarks demonstrate that TBDQ-Net achieves a favorable efficiency-accuracy trade-off in challenging scenarios. Specifically, TBDQ-Net outperforms leading TBD methods by 6.0 IDF1 points on DanceTrack and achieves the best performance among TBQ methods in the crowded MOT20 benchmark. Relative to MOTRv2, TBDQ-Net reduces trainable parameters by approximately 80\% while accelerating practical inference by 37.5\%.
These results highlight TBDQ-Net as an efficient alternative to heavy architectures, showcasing the efficacy of lightweight design.
Source code is publicly available at  \url{https://github.com/FaithFlow/TBDQ-Net}.
\end{abstract}


\end{frontmatter}

\footnotetext[1]{\textsuperscript{*}Corresponding author: Xiaobo Lu, email: xblu2013@126.com}

\section{Introduction}
Multi-object tracking (MOT) is a pivotal task in applications involving scene understanding \cite{li2025indoor}, multi-drone tracking \cite{liu2023robust}, etc. Given a video with target classes, the goal is to recognize, localize and assign consistent identification numbers to objects over time. It essentially consists of two tasks: detection and association. Handling the relationship between them has been a central theme in the field. Accordingly, there are two dominant paradigms: the conventional tracking-by-detection (TBD) \cite{bewley2016simple, zhang2022bytetrack, mandel2023detection} and the emerging tracking-by-query (TBQ) \cite{zeng2022motr, gao2023memotr, yan2025comot}, which have complementary merits and demerits.

The TBD paradigm employs a decoupled framework, first detecting objects spatially and then associating them temporally. The modular framework facilitates the utilization of pretrained, high-performance detectors \cite{zhou2020tracking, ge2021yolox} and lightweight association algorithms \cite{bewley2016simple, zhang2022bytetrack}, leading to outstanding practical efficiency. However, in the typical association process, appearance features and motion patterns are extracted independently and integrated heuristically, \textit{resulting in a fragmented pipeline that lacks coherent modeling capabilities in complex scenarios}. Adopting a different philosophy, the TBQ paradigm employs an end-to-end framework to effectively address these limitations. It substantially enhances association capabilities through coherent utilization of tracking data. Nonetheless, TBQ methods typically couple detection and association, \textit{meaning that their powerful association performance is intrinsically tied to joint training throughout the model.} This tight coupling often leads to task competition \cite{zeng2022motr, yan2025comot} and substantially impedes overall efficiency, resulting in high training barriers and computational costs. In essence, despite their individual merits, these two mainstream paradigms possess complementary advantages: TBD excels in efficiency but struggles with comprehensive contextual understanding, while TBQ provides robust and expressive modeling capabilities, but at a significant computational cost. This clear structural conflict highlights a pressing demand for their unification to achieve both efficiency and robust association.

\begin{figure}
    \centering
    \setlength{\belowcaptionskip}{-0.3cm}
    \setlength{\abovecaptionskip}{-0.05pt}
    \includegraphics[width=1\linewidth]{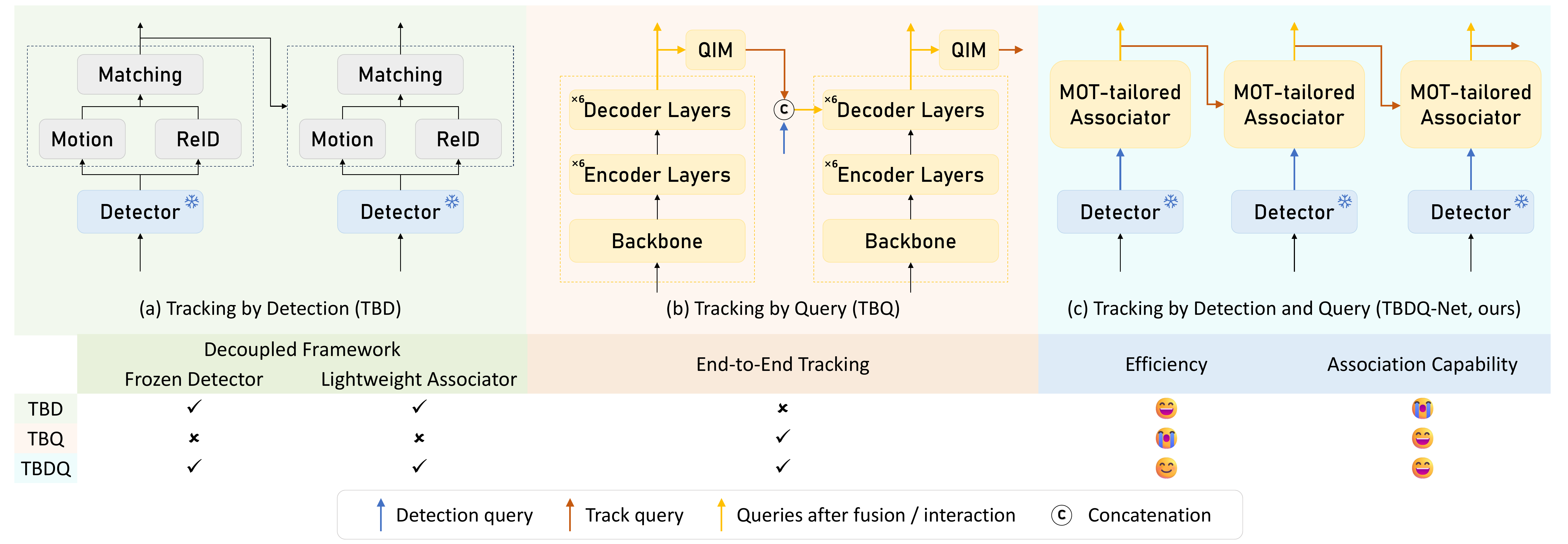}
    \caption{\textbf{Comparison of mainstream paradigms.} (a) Tracking-by-Detection (TBD): A decoupled but fragmented framework that typically relies on heuristic, non-differentiable matching, limiting association robustness in complex scenarios. (b) Tracking-by-Query (TBQ): An end-to-end trainable framework that enhances coherence but suffers from heavy computational costs due to the coupled detection-association architecture. (c) Tracking-by-Detection-and-Query (TBDQ-Net, ours): A unified framework integrating a frozen, pre-trained detector with a lightweight, learnable associator. It effectively synthesizes the advantages of both paradigms: achieving the high efficiency of decoupled architectures while maintaining the strong end-to-end tracking capability.
    }
    \label{fig:cmp_all}
\end{figure}

Recognizing this demand, recent pioneers such as MOTIP \cite{gao2025multiple} and MOTRv2 \cite{zhang2023motrv2} have explored hybrid directions that combine the merits of both paradigms. Specifically, MOTIP transforms heuristic association into an end-to-end ID prediction task, while MOTRv2 leverages a frozen YOLOX to guide a query-based tracker. Although these methods present significant progress, they inherently rely on joint optimization or heavy computation for best performance.
For instance, MOTIP necessitates the joint training of the detector and the ID predictor. And freezing the detector leads to a notable performance drop, indicating a dependency on co-adapted features. Similarly, MOTRv2 retains a heavy Transformer architecture for feature refinement, which limits its potential for efficiency. These observations suggest that achieving robust, end-to-end tracking within a truely decoupled, lightweight framework remains a non-trivial challenge, requiring careful architectural design beyond simple integration.

In this work, we further develop this unification trend by proposing the tracking-by-detection-and-query framework, TBDQ-Net.
Unlike previous works that merely use the detector for proposal generation, TBDQ-Net fully leverages the pre-trained detector as a source of both high-quality object representations and rich scene context (e.g., backbone features). By integrating this frozen detector with a well-structured lightweight associator, we ensure the intrinsic efficiency of the framework. Within this streamlined architecture, we achieve highly competitive tracking performance by addressing two critical challenges:
(1) Dual-stream information interaction. The Basic Information Interaction (BII) module is proposed to facilitate bidirectional communication. The detection stream (BII-D) incorporates track information to suppress potential conflicts, while the track stream (BII-T) integrates current observations and history memory to maintain temporal stability against disturbances and occlusions.
(2) Spatiotemporal alignment. Following the semantic updates in BII, we introduce the Content-Position Alignment (CPA) module. It simultaneously refines both content and positional components, bridging the semantic-spatial gap and delivering well-aligned representations for the final decoding.
Ultimately, TBDQ-Net sets a new performance reference by delivering unparalleled efficiency and robust tracking performance, effectively unifying the primary advantages of both TBD and TBQ. As visually summarized in Figure \ref{fig:cmp_all}, our framework uniquely marries the strong end-to-end tracking capability with the high structural efficiency of a lightweight, decoupled architecture.

We extensively evaluate TBDQ-Net on three challenging datasets, including the large-scale DanceTrack \cite{sun2022dancetrack} and SportsMOT \cite{cui2023sportsmot}, as well as the extremely dense MOT20 \cite{dendorfer2020mot20} benchmark.
Compared to TBD methods, our framework demonstrates superior association robustness in challenging scenarios. For instance, on DanceTrack, TBDQ-Net surpasses the state-of-the-art Hybrid-SORT \cite{yang2024hybrid} by +3.7 HOTA and +6.0 IDF1, highlighting its capability to handle complex non-linear motion patterns.
Compared to TBQ methods, TBDQ-Net significantly accelerates both training and inference while maintaining competitive accuracy. Notably, it achieves this with 80\% fewer parameters and 37.5\% faster practical inference than MOTRv2, demonstrating that high performance can be achieved with lightweight architectures.
Overall, our contributions are summarized as follows:

\begin{itemize}

\item We propose TBDQ-Net, a streamlined framework that advances the combination of TBD and TBQ paradigms, effectively resolving the inherent efficiency-accuracy trade-off in MOT. By demonstrating that a lightweight associator can achieve high performance with a frozen detector, TBDQ-Net offers a compelling alternative to the prevailing trend of heavy modules and joint training in the conventional query-based end-to-end tracking.

\item We develop a novel learnable associator tailored for this framework. It features the BII module for effective dual-stream interaction and the CPA module for precise information alignment, enabling the system to handle complex scenarios.

\item We demonstrate TBDQ-Net's effectiveness and practical utility through extensive experiments across challenging real-world scenarios, including dance, sports, and dense street scenes. It achieves a highly remarkable balance between tracking efficiency and accuracy, establishing a new benchmark in the field.

\end{itemize}

\section{Related Work}

Significant advancements in multi-object tracking (MOT) have been driven by the evolution of two primary paradigms: tracking-by-detection (TBD) and tracking-by-query (TBQ). Recently, a third direction has emerged, aiming to unify their strengths. This section reviews these approaches and positions our work within this domain.

\subsection{Tracking by Detection} 
This paradigm has long dominated the field due to its modular efficiency. Early works like SORT  \cite{bewley2016simple} established the baseline by combining Kalman filter with heuristic matching. Subsequent methods have focused on refining specific components: FairMOT \cite{zhang2021fairmot} explores joint detection and embedding learning; ByteTrack \cite{zhang2022bytetrack} enhances association by utilizing low-confidence detections; OC-SORT \cite{cao2023observation} improves motion modeling for non-linear trajectories; FineTrack \cite{ren2023focus} explores diverse fine-grained representations for discriminative target identification.  More recently, PD-SORT \cite{wang2025pd} integrates pseudo-depth cues into the Kalman filter and matching cost to robustly handle occlusions. Furthermore, ReTrackVLM \cite{bayraktar2025retrackvlm} utilizes VLM embeddings for zero-shot re-identification, while OVSORT \cite{li2025open} addresses open-vocabulary challenges through adaptive normalization and joint motion-appearance modeling. SeaTrack \cite{ding2025seatrack} addresses nearshore MOT mainly based on motion information. Despite their efficiency, TBD methods fundamentally rely on a fragmented pipeline and heuristic matching algorithms. This separation often leads to sub-optimal use of temporal context, limiting their association robustness in complex scenarios.

\subsection{Tracking by Query} 
Emerging with the rise of Transformers, TBQ methods formulate tracking as a set prediction problem, achieving superior association robustness through end-to-end learning.
Representative of this paradigm, TrackFormer \cite{meinhardt2022trackformer} and MOTR \cite{zeng2022motr} concurrently explored the use of track queries to bridge temporal information. Specifically, TrackFormer \cite{meinhardt2022trackformer} treats tracking as an auto-regressive process across adjacent frames, while MOTR \cite{zeng2022motr} introduces a recursive update and propagation mechanism for track queries, effectively serving as implicit memory carriers for object states. In contrast to this recursive approach, MO3TR \cite{zhu2022looking} explicitly models long-term dependencies by maintaining a history of object embeddings and employing a separate temporal Transformer for state prediction. Other works extend these foundations: MeMOT \cite{cai2022memot} and MeMOTR \cite{gao2023memotr} enhance long-term memory with an external memory bank; CO-MOT \cite{yan2025comot} leverages additional supervision to alleviate task competition.
While these methods eliminate heuristic matching, they typically rely on a heavy, coupled Transformer architecture to perform detection and association simultaneously. This leads to significant computational burdens and task competition between detection and association, making the models difficult to train and slow to infer.

\subsection{Emerging Trend}
Recognizing the complementary strengths of TBD in efficiency and TBQ in robustness, a new trend has emerged to decouple detection from association while maintaining end-to-end tracking.
MOTIP \cite{gao2025multiple} attempts this by transforming the association of TBD into an end-to-end ID prediction task. However, it still relies on joint training of the detector and the heavy ID predictor. As shown in their studies, freezing the detector or reducing ID decoder layers could result in a significant performance drop.
MOTRv2 \cite{zhang2023motrv2} decouples detection by using a frozen YOLOX \cite{ge2021yolox} to guide the MOTR architecture. Although this alleviates the detection-association conflict, it retains the computationally expensive DETR encoder-decoder structure for feature refinement, thus offering limited improvements in overall efficiency.
Distinct from these attempts, TBDQ-Net represents a more thorough realization of this decoupled paradigm. We demonstrate that a properly designed lightweight associator can yield state-of-the-art performance even with a strictly frozen detector, validating the feasibility of high-performance tracking without heavy architectural redundancy.

\section{Methodology}

\begin{figure}[t]
    \centering
    \includegraphics[width=1.0\linewidth]{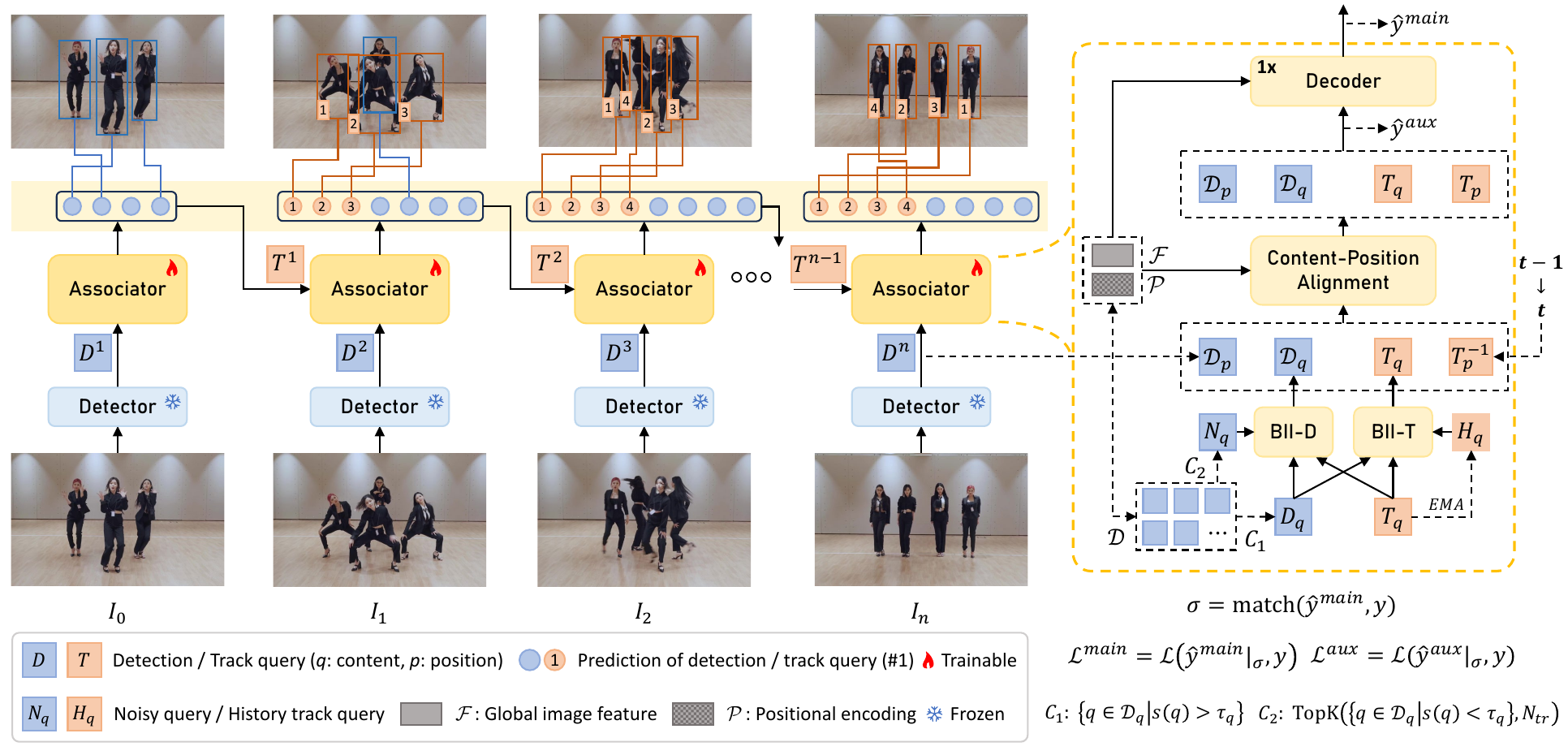}
    \caption{
    \textbf{Architectural overview of TBDQ-Net.} The framework follows an end-to-end query-based tracking paradigm, modeling target trajectories via track queries ($T$), while capturing emerging objects through detection queries ($D$). For each frame, a frozen detector extracts a raw candidate pool $\mathcal{D}$. Subsequently, the selection strategies $C_1$ and $C_2$ partition the raw candidates into qualified detection queries $D_q$ and noisy queries $N_q$ based on confidence scores $s(q)$ and the threshold value $\tau_q$. The internal mechanism of the associator (dashed box) primarily consists of two key components:
    (1) The Basic Information Interaction (BII) module. It facilitates dual-stream refinement: the BII-D stream filters detection queries by perceiving existing tracks to alleviate task conflicts, while the BII-T stream updates trajectories by integrating historical context ($H_q$) with the latest detection priors ($D$).
    (2) The Content-Position Alignment (CPA) module. It reconciles the semantic gap, which is primarily indicated by the temporally lagged $T_p^{-1}$, between query content ($Q_q$) and spatial embeddings ($Q_p$) using global image features $\mathcal{F}$ and positional encodings $\mathcal{P}$.
    To enhance training, hierarchical supervision is applied through auxiliary ($\hat{y}^{aux}$) and main ($\hat{y}^{main}$) predictions.
    }
    \label{fig:overall}
    \vspace{-10pt}
\end{figure}

\subsection{Overview}

The pipeline of TBDQ-Net is illustrated in Figure \ref{fig:overall}. Following the end-to-end query-based tracking paradigm, the framework models target trajectories through track queries, while detection queries are responsible for capturing newly emerging objects.
At the heart of this framework, the Basic Information Interaction (BII) module facilitates dual-stream communication between detection and track queries. Specifically, the BII-D stream allows detection queries to perceive existing tracks, suppressing redundant predictions and alleviating task conflicts. Simultaneously, the BII-T stream updates track queries with the current detection priors for immediate information update, while integrating historical context to ensure temporal stability. To bridge the potential semantic gap between query content and positional embeddings induced by the interaction process, we further introduce the Content-Position Alignment (CPA) module. This ensures spatio-temporal consistency in the coherent optimization of query components. By addressing these fundamental MOT challenges, TBDQ-Net successfully achieves a superior balance between computational efficiency and robust association in a lightweight end-to-end framework.

\subsection{Problem Formulation}

Given a video sequence $\mathcal{V} = \{I_t\}_{t=1}^L$, TBDQ-Net aims to produce a trajectory set through the end-to-end optimization of track queries. At each moment $t$, a frozen pre-trained detector (e.g., YOLOX) extracts high-confidence object candidates from image $I_t$. These candidates are transformed into detection queries $D_q^{t}$, representing potential new objects. Meanwhile, track queries $T_q^{t-1}$ propagate from the last moment $t-1$, linking to existing objects. Motivated by the Transformer-based object detection \cite{liu2022dabdetr, zhao2024detrs}, each query $Q \in \{D, T\}$ is decoupled into two essential components: $Q = \{Q_q, Q_p\}$, where $Q_q \in \mathbb{R}^C$ denotes the content part that focuses on the semantic information, and $Q_p \in \mathbb{R}^C$ signifies the position part that represents the localization information. Following the standard operation, $Q_p$ is produced by the sinusoidal encoding \cite{liu2022dabdetr} of the target bounding boxes $Q_b \in \mathbb{R}^4$. Additionally, in the detection process, the global image features $\mathcal{F}$ and positional encoding $\mathcal{P}$ are easily acquired. Given these embeddings, the goal of the associator is to facilitate information interaction for $D_q^t$ and $T_q^{t-1}$ through the BII modules, and temporal alignment for $Q_q$ and $Q_p$ in the CPA module. Finally, the entire associator is optimized end-to-end through a hierarchical supervision mechanism.

\subsection{Integration of Pretrained Detectors}

Our associator is designed to be detector-agnostic, facilitating seamless integration with both CNN-based and Transformer-based architectures. Beyond the standard detection outputs, such as bounding boxes ($D_b$) and confidence scores ($s$), the associator only requires three components derived from the detector's inference process: global image features ($\mathcal{F}$), positional encodings of the backbone feature map ($\mathcal{P}$), and object representation embeddings ($D_q$).

For CNN-based Detectors (e.g., YOLOX), the global image features $\mathcal{F}$ are aggregated by flattening the multi-scale features from the feature pyramid network (FPN), while their corresponding spatial positional encodings $\mathcal{P}$ are synchronously acquired. These representations are readily accessible during the detector's inference stage without significant overhead. The object representation embeddings $D_q$ are obtained by sampling the FPN feature maps based on the predicted bounding boxes $D_b$.

For Transformer-based Detectors (e.g., DAB-DETR), these models are inherently compatible with our framework. Global features $\mathcal{F}$ and their corresponding positional encodings $\mathcal{P}$ are obtained from the refinement features of the encoder. Meanwhile, the object representation embeddings $D_q$ are derived directly from the output embeddings of the last decoder layer.

By decoupling backbone feature refinement from association, TBDQ-Net avoids the expensive joint training typical of conventional TBQ models while maintaining high flexibility across various detection models.

\subsection{Basic Information Interaction Module}
\label{sec:bii}

As the core component of the learnable associator, the Basic Information Interaction (BII) module is designed to facilitate a holistic, bidirectional information flow between detection and track queries. This design serves a twofold objective: detection queries related to existing objects are suppressed by track queries to mitigate potential conflicts; simultaneously, track queries are updated by integrating new detection information to ensure their temporal validity. Technically, the BII module is implemented through the scaled dot-product attention mechanism \cite{vaswani2017attention}:
\begin{equation}
\label{bii}
\begin{aligned}
O_1&=\operatorname{softmax}\left(\frac{Q K^T}{\sqrt{d}}\right) \cdot V_1, \\ 
O_2&=\operatorname{norm}\left(O_1+V_2\right), \\ 
O_3&=\operatorname{norm}\left(\operatorname{FFN}\left(O_2\right)+O_2\right).
\end{aligned}
\end{equation}
Specifically, the standard vector Value ($V$) is replaced by two distinct vectors $V_1$ and $V_2$, where $V_1$ serves as the source of information to update the base features contained in $V_2$. $d$ is the dimension of vectors $Q$, $K$, $V_1$, and $V_2$. $\operatorname{FFN}$ represents the standard feed-forward neural network \cite{vaswani2017attention}. To facilitate spatial-aware interaction, we define the unified embedding $\widetilde{Q} = Q_q + Q_p$, and $Q_p = \text{PE}(Q_b)$, where $\text{PE}(\cdot)$ denotes the sinusoidal positional encoding \cite{liu2022dabdetr}. This representation is utilized for Query ($Q$) and Key ($K$) formulations to compute attention weights, while the Value ($V$) remains as the pure content $Q_q$ to preserve semantic integrity during the update. On top of this mechanism, the bidirectional interaction is realized through specific configurations of $Q, K, V_1$, and $V_2$, formulating the BII-D and BII-T streams. 

Specifically, detection queries are updated in the BII-D stream:
\begin{equation}
\label{biidet}
\begin{aligned}
&Q=\widetilde{D_q}, K=\operatorname{concat}(\widetilde{D_q},\widetilde{T_q}), \\ 
&V_1=\operatorname{concat}(D_q,N_q), V_2=D_q,
\end{aligned}
\end{equation}
where $\text{concat}(\cdot)$ denotes the concatenation along the object dimension, preserving the channel dimension $C$. To reduce the impact of background noise and improve interaction quality, we employ a confidence-based partitioning strategy for detection candidates $\mathcal{D}_q$. Specifically, we utilize the strategy $C_1: \{q \in \mathcal{D}_q \mid s(q) > \tau_q\}$ to filter out low-quality candidates, retaining only high-confidence queries $D_q$ before interaction. Furthermore, to resolve potential conflicts where a detection query spatially overlaps with an existing track, we introduce noisy queries $N_q$ as active distractors. They are constructed via the strategy $C_2: \text{TopK}(\{q \in \mathcal{D}_q \mid s(q) < \tau_q\}, N_{tr})$, selecting the top-$N_{tr}$ highest-scoring queries from the filtered low-confidence set. These queries typically correspond to redundant detections or background clutter near objects, serving as hard negative samples. Through the attention mechanism, the network learns to assign higher attention weights to these noisy queries when a detection query is similar to a track query, thereby disrupting its feature aggregation and effectively alleviating inherent conflicts.

Track queries are updated in the BII-T path:
\begin{equation}
\label{biitrack}
\begin{aligned}
&Q=\widetilde{T_q}, K=\operatorname{concat}(\widetilde{D_q},\widetilde{H_q}), \\ 
&V_{1}=\operatorname{concat}(D_q,H_q), V_{2}=T_q.
\end{aligned}
\end{equation}
In this process, $D_q$ represents the immediate observations to ensure the continuity of tracking and adapt to evolving target states, which effectively prevents tracks from drifting or terminating during normal movement. $H_q$ serves as long-term memory to handle challenging scenarios where objects are temporarily occluded or blurred. $H_q$ is collected by the exponential moving average (EMA) method:
\begin{equation}
\label{ema}
{H_q^t} =
\begin{cases}
w*T_q^t+(1-w)*H_q^{t-1},&{\text{if}}\ t>0,\\
{\emptyset,}&{\text{if}}\ t=0,
\end{cases}
\end{equation}
where $w$ indicates the update weight of new information. The position part of $\widetilde{H_q}$ is identical to $\widetilde{T_q}$, i.e., $H_p = T_p$, which is spatially synchronized with the current frame.  

Note that only the qualified detection queries $D_q$ are updated by the BII module, while the original features are preserved for the rest of the candidates to maintain spatial context before entering the subsequent alignment module.

\subsection{Content-Position Alignment Module}
The BII module focuses on the interaction of the content components $Q_q$ of queries. Consequently, the corresponding positional component $Q_p$ remains unchanged. Especially for track queries, their positional component $T_p$ is attached to the spatial coordinates inherited from the previous moment $t-1$. This leads to a spatio-temporal misalignment that inevitably degrades the reliability of the associator. To bridge this gap, we introduce the Content-Position Alignment (CPA) module to dynamically synchronize the query's spatial information with its semantic representation. Formally, given the updated content $Q^u_q$ and the stagnant position $Q^{-1}_p$, the aligned content and position part $Q^a_q$, $Q^a_p$ are computed as:
\begin{equation}
\label{cpa}
Q^a_q, Q^a_p = f(Q^u_q, Q^{-1}_p),
\end{equation}
where $f(\cdot)$ represents a learnable transformation. We implement this function via the modulated cross-attention mechanism \cite{liu2022dabdetr}, leveraging its intrinsic capacity to refine positional and object embeddings in a unified manner. Specifically, this design modulates semantic feature representations with explicit bounding box priors, facilitating a deep integration of spatial geometry and content information. By encoding the localization awareness directly into the refinement process, the CPA module effectively bridges the gap between the updated content and its corresponding spatial state.

Furthermore, to fortify the core efficacy of the associator, we apply an auxiliary supervision to the outputs of the CPA module. Enforcing optimization at this intermediate stage ensures that the output query is a well-calibrated representation for subsequent decoding. Details about this auxiliary supervision are illustrated in Section \ref{sec:traininginference}.

\subsection{Association Decoder}

Through the BII and CPA modules, detection and track queries undergo thorough semantic exchange and information alignment across their content and positional components. Subsequently, these queries are directly decoded into predictions using a single Transformer decoder layer.
This architecture preserves the core philosophy of end-to-end tracking: modeling the full trajectory lifecycle through the continuous optimization of track queries.
In contrast to traditional TBD matching strategies that rely on heuristic, non-differentiable post-processing (e.g., IoU matching), this learnable query propagation enables the model to adaptively handle complex motion patterns and occlusions.
Therefore, even with a frozen detector, TBDQ-Net maintains the end-to-end nature, effectively integrating robust detection priors with differentiable tracking policies.

\subsection{Training and Inference}
\label{sec:traininginference}

Although TBDQ-Net introduces a novel associator architecture, it adheres to the mainstream query-based tracking paradigm for prediction logic. The network outputs predictions through two sets of queries: detection queries are responsible for new-born objects, while track queries link to existing tracks.

In the training stage, we employ a bipartite matching strategy for label assignment. For new-born objects at frame $I_t$, we perform bipartite matching between the predictions of detection queries $\hat{y}_{det}^t$ and the ground truth of new objects $y_{new}^t$ to obtain $\sigma_{det}^t$:
\begin{equation}
    \sigma_{det}^t = \arg\min_{\hat{\sigma} \in \Omega_t} \mathcal{L}_{match}(\hat{y}_{det}^t |_{\hat{\sigma}, y_{new}^t),}
\end{equation}
where $\mathcal{L}_{match}$ is the matching cost composed of classification and regression losses \cite{liu2022dabdetr}, and $\Omega_t$ denotes the space of all possible bipartite matches between the detection queries and the new-born objects. To ensure temporal consistency, the assignment set is updated recursively. Specifically, the complete set of tracks is represented by the assignment $\sigma_{tr}^t$, which is formed by the union of the assignments inherited from the previous frame ($\sigma_{tr}^{t-1}$) and the newly identified objects in the current frame ($\sigma_{det}^t$):
\begin{equation}
    \sigma_{tr}^t = \begin{cases} \sigma_{tr}^{t-1} \cup \sigma_{det}^t, & \text{if } t > 0, \\ \emptyset, & \text{if } t = 0. \end{cases}
\end{equation}
Note that this assignment is derived solely from the main predictions $\hat{y}^{main}$ and is then used to calculate the loss function for both main and auxiliary predictions:
\begin{equation}
    \mathcal{L}_{total} = \mathcal{L}(\hat{y}^{main}|_{\sigma}, y) + \mathcal{L}(\hat{y}^{aux}|_{\sigma}, y).
\end{equation}
Specifically, the individual loss items follow the standard DETR-style formulation \cite{liu2022dabdetr}:
\begin{equation}
    \mathcal{L}(\hat{y}, y) = \lambda_{cls}\mathcal{L}_{cls} + \lambda_{L1}\mathcal{L}_{L1} + \lambda_{giou}\mathcal{L}_{giou},
\end{equation}
where $\mathcal{L}_{cls}$ is the focal loss for classification, and $\mathcal{L}_{L1}, \mathcal{L}_{giou}$ are the L1 and GIoU losses for bounding box regression. Consistent with the standard configuration in MOTR  \cite{zeng2022motr}, we set the loss balancing coefficients as $\lambda_{cls}=2, \lambda_{L1}=5, \lambda_{giou}=2$.

During inference, the framework tracks objects in an online manner. Detection queries with prediction scores above $\tau_n$ initiate new tracks. Existing tracks are updated if their corresponding track query predictions exceed $\tau_n$. Otherwise, they are marked as the inactive state. Inactive tracks will be removed if they remain lost for more than $L_{max}$ consecutive frames.

For a detailed overview of the training and inference procedures, we provide the pseudo-code in Algorithm \ref{alg:pipeline}.

{ 
    \footnotesize  
    \setlength{\baselineskip}{1.2em} 
    \setlength{\parindent}{0pt} 

    \rule{\linewidth}{1.2pt} 
    
    \vspace{-1.2em} 
    
    \captionof{algorithm}{\textbf{Training and Inference Pipeline of TBDQ-Net.}}
    \label{alg:pipeline}
    
    \vspace{-0.8em} 
    
    \rule{\linewidth}{0.6pt} 
    
    \vspace{0.2em} 

    \begin{algorithmic}[1]
        \REQUIRE Video sequence $\mathcal{V} = \{I_t\}_{t=1}^L$, Pretrained Detector $\text{Det}(\cdot)$, Association Modules: $\text{BII-D/T}(\cdot)$, $\text{CPA}(\cdot)$, $\text{Decoder}(\cdot)$, Prediction Head $\text{Head}_\text{main/aux}(\cdot)$; Hyperparameters: score threshold values $\tau_q, \tau_n$, max lost frames $L_{\text{max}}$.
        \ENSURE Tracklets $\mathcal{T}$.
        
        \STATE Initialize history track queries $H_q \leftarrow \emptyset$, tracklets $\mathcal{T} \leftarrow \emptyset$, track assignments $\sigma_{tr}^0 \leftarrow \emptyset$.
        
        \FOR{$t = 1$ to $L$}
            \STATE \textbf{// 1. Feature Extraction \& Candidate Generation}
            \STATE $\mathcal{F}, \mathcal{P}, \mathcal{D}_q, \mathcal{D}_p, s \leftarrow \text{Det}(I_t)$ \MyComment{Frozen detector} 
            
            \STATE \textbf{// 2. Query Filtering \& Preparation}
            \STATE $D_q, D_p \leftarrow \{q \in \mathcal{D}_q, p \in \mathcal{D}_p \mid s(q) > \tau_q\}$ \MyComment{Selection strategy $C_1$} 
            \STATE $T_q \leftarrow \mathcal{T}.\text{query}, \quad T_p \leftarrow \mathcal{T}.\text{box}$
            \STATE $H_q \leftarrow \text{EMA}(T_q, H_q)$ 
            \STATE $N_{tr} \leftarrow |\mathcal{T}|$
            
            \STATE \textbf{// 3. Association \& Aux. Supervision}
            \IF{$N_{tr} > 0$}
                \STATE $N_q \leftarrow \text{TopK}(\{q \in \mathcal{D}_q \mid s(q) < \tau_q\}, N_{tr})$ \MyComment{Selection strategy $C_2$} 
                \STATE $D_q \leftarrow \text{BII-D}(T_q, D_q, T_p, D_p, N_q)$
                \STATE $T_q \leftarrow \text{BII-T}(T_q, D_q, T_p, D_p, H_q)$
                \STATE $Q_q \leftarrow \text{Concat}(T_q, D_q), \quad Q_p \leftarrow \text{Concat}(T_p, D_p)$
                \STATE $Q_q, Q_p \leftarrow \text{CPA}(\mathcal{F}, \mathcal{P}, Q_q, Q_p)$
                \STATE $\hat{y}^{\text{aux}} \leftarrow \text{Head}_{\text{aux}}(Q_q, Q_p)$ 
            \ELSE
                \STATE $Q_q \leftarrow D_q, \quad Q_p \leftarrow D_p, \quad \hat{y}^{\text{aux}} \leftarrow \emptyset, \quad \mathcal{L}_{\text{aux}} \leftarrow 0$
            \ENDIF
            
            \STATE \textbf{// 4. Final Alignment \& Prediction}
            \STATE $Q_{q}, Q_p \leftarrow \text{Decoder}(\mathcal{F}, \mathcal{P}, Q_q, Q_p)$
            \STATE $\hat{y}^{\text{main}} \leftarrow \text{Head}_{\text{main}}(Q_{q}, Q_p)$
            
            \STATE \textbf{// 5. Track Management for Training / Inference}
            \STATE $\hat{y}_{\text{track}}^t \leftarrow \hat{y}^{\text{main}}[1 \dots N_{tr}], \quad \hat{y}_{\text{det}}^t \leftarrow \hat{y}^{\text{main}}[N_{tr}+1 \dots \text{end}]$
            \IF{training}
                \STATE $\sigma_{det}^t \leftarrow \text{Match}(\hat{y}_{\text{det}}^t, y_{new}^t)$ \MyComment{Match with main predictions} 
                \STATE $\sigma \leftarrow\sigma_{tr}^{t-1} \cup \sigma_{det}^{t}$
    
                \STATE $\mathcal{L}_{\text{main}} \leftarrow \text{Loss}(\hat{y}^{\text{main}}|_{\sigma}, y)$ \MyComment{Share assignment for $\mathcal{L}^{\text{main}}, \mathcal{L}^{\text{aux}}$}
                \STATE $ \mathcal{L}_{\text{aux}} \leftarrow \text{Loss}(\hat{y}^{\text{aux}}|_{\sigma}, y) \quad (\text{if}  \; \hat{y}^{\text{aux}} \neq \emptyset)$  
                \STATE $\mathcal{L}^{\text{total}} \leftarrow \mathcal{L}^{\text{main}} + \mathcal{L}^{\text{aux}}$
                \STATE $\mathcal{T} \leftarrow \{ \text{Track}(\hat{y}_{main}(i)) \mid i \in \sigma \}$ \MyComment{Update tracklets} 
                \STATE $\sigma_{tr}^t \leftarrow \sigma$
                
            \ELSE
                \STATE $\mathcal{T} \leftarrow \{ \text{Track}(\hat{y}) \mid \hat{y} \in \hat{y}_{track}\} $ \MyComment{Update all previous tracks}
                \STATE $\mathcal{T}_{new} \leftarrow \{ \text{Track}(\hat{y}) \mid \hat{y} \in \hat{y}_{det}, \hat{y}.\text{score} > \tau_n \}$ \MyComment{Initialize new tracks} 
                \STATE $\mathcal{T}_{lost} \leftarrow \{ \tau \in \mathcal{T} \mid \tau.\text{score} < \tau_n \}$ \MyComment{Handle inactive tracks}
                \STATE $\forall \tau \in \mathcal{T}_{lost}: \tau.\text{lost} \mathrel{+}= 1$
                \STATE $\mathcal{T}_{delete} \leftarrow \{ \tau \in \mathcal{T} \mid \tau.lost > L_{max} \}$ \MyComment{Remove dead tracks}
                \STATE $\mathcal{T} \leftarrow (\mathcal{T} \cup \mathcal{T}_{new}) \setminus \mathcal{T}_{delete}$  \MyComment{Update tracklets}
            \ENDIF
        \ENDFOR
    \end{algorithmic}
    
    \noindent \rule{\linewidth}{1.2pt} 
}

\section{Experiments}
\subsection{Datasets}
To thoroughly evaluate the effectiveness of TBDQ-Net, we choose datasets that represent the most challenging scenarios in modern MOT. Specifically, we select DanceTrack \cite{sun2022dancetrack} and SportsMOT \cite{cui2023sportsmot} to assess the model's capability in handling highly non-linear, fast-moving objects. These dynamic environments pose tremendous difficulties for traditional motion priors, serving as ideal benchmarks for our learnable associator.
In addition, we include MOT20 \cite{dendorfer2020mot20} to evaluate performance in extremely crowded scenes. This tests the capacity to maintain identity consistency against severe spatial overlap and frequent occlusions.
By covering this comprehensive scope, which ranges from sparse, highly-dynamic motion to ultra-dense, complex interaction, we aim to validate the robustness and generalization of our approach across distinct motion modalities and scene densities. Detailed statistics for these datasets are listed in Table \ref{tab:dataset_statistics}.

\begin{table}[t]
\caption{\textbf{Detailed statistics of the related datasets.} The number of frames (K) / sequences in training, validation and test subsets for each dataset. Density denotes the average number of boxes per frame.}
\label{tab:dataset_statistics}
\footnotesize
\setlength{\tabcolsep}{2.6pt}
\centering
\begin{tabular}{c|cccc|c}
\toprule
Datasets   & Training & Validation  & Test     & Total   & Density   \\ \midrule
DanceTrack \cite{sun2022dancetrack} & 41.8 / 40  & 25.5 / 25     & 38.6 / 35  & 105.9 / 100  &  5.4  \\
SportsMOT \cite{cui2023sportsmot}  & 28.6 / 45  & 27.0 / 45     & 94.8 / 150 & 150.4 / 240  &  10.8  \\
MOT20 \cite{dendorfer2020mot20}      & 8.9 /  4    & --            & 4.5  / 4    & 13.4 / 8    &  123.2  \\  \bottomrule 
\end{tabular}
\vspace{-8pt}
\end{table}

\vspace{-4pt}
\subsection{Metrics}
To ensure a fair and standardized comparison with existing literature, we report tracking performance on DanceTrack and SportsMOT using the five most widely adopted metrics: HOTA \cite{luiten2021hota}, DetA, AssA, MOTA \cite{bernardin2008evaluating} and IDF1 \cite{ristani2016performance}. HOTA is a balanced metric, comprehensively evaluating tracking performance containing both detection and association, and $HOTA = \sqrt{DetA \cdot AssA}$. DetA and MOTA emphasize detection performance, while AssA and IDF1 focus on association capabilities. For the MOT20 benchmark, we adhere to its specific evaluation protocol by additionally reporting FP, FN, and IDSW, which are standard indicators for this dataset. Specifically, $\text{MOTA} = 1 - (\text{FP} + \text{FN} + \text{IDSW}) / {\text{GT}}$. FP, FN, IDSW and GT represent the number of predicted false positives, false negatives, ID switches and annotated ground truth boxes, respectively.
\subsection{Implementation Details}

\textit{General Details:} We adopt an initial learning rate of $1.2\times10^{-4}$ for all datasets. On DanceTrack, the training schedule is (6, 10, 12), meaning that the model is trained for 12 epochs, with the initial learning rate dropped by 10 at the 6th and 10th epoch, respectively. To accommodate the smaller data volumes in SportsMOT and MOT20, the training schedule expands to (12, 16, 20) and (20, 25, 30), respectively. For a fair comparison, TBDQ-Net utilizes the publicly released YOLOX model pretrained by ByteTrack \cite{zhang2022bytetrack}. Unless particularly specified, the image processing and sampling strategies are identical to those of CO-MOT \cite{yan2025comot} and MOTRv2 \cite{zhang2023motrv2}. In all experiments, hyperparameters $\tau_q$ (Section \ref{sec:bii}), $\tau_n$ and $L_{max}$ (Section \ref{sec:traininginference}) are set to 0.3, 0.5 and 20, respectively. This configuration serves as a robust baseline across diverse scenarios. Empirical analysis suggests that while these values are generally near-optimal, they can be further adapted based on scene-specific attributes. \textit{To focus on the essence of tracking and avoid artifacts from unrealistic, clumsy motion patterns \cite{gao2025multiple}, we did not utilize any extra data for joint training.}

\textit{Ablation Studies:} The major ablation study is conducted in DanceTrack, due to the sufficient and high-quality data in training and validation subsets. For fast iteration, TBDQ-Net incorporates a pretrained DAB-DETR as the detector and is trained in lower resolutions. The shortest side ranges from 480 to 800 pixels with a step of 32 pixels, and the longest side is at most 1,333 pixels.

\textit{MOT20:} Given that the MOT20 training set contains significantly fewer frames (3x to 6x less) than DanceTrack and SportsMOT, training data-hungry query-based models poses a risk of overfitting. To mitigate this data scarcity and ensure effective learning, we adopt two standard augmentation strategies following MOTR \cite{zeng2022motr}: (1) We increase the training clip length to capture more coherent long-term temporal cues from the limited sequences. (2) We introduce random insertion of negative track instances with probability $p_i$, and random dropout of positive track instances with probability $p_d$, to artificially enrich the diversity of tracking scenarios.
Specific configurations for these parameters are provided in the ablation study. As there is no validation set in MOT20, we follow CenterTrack \cite{zhou2020tracking} to divide each sequence in the training set into two parts: the first part is used for training, and the second part serves as the validation set.

It is worth noting that apart from these necessary adjustments to handle severe data limitation, all other core configurations remain consistent with those used in DanceTrack and SportsMOT, demonstrating the framework's fundamental stability.

\subsection{Comparison with Other Methods}
\label{sec:cmp_sota}

Initially, we compare the \textit{efficiency} of TBDQ-Net against representative TBQ methods, demonstrating its superiority in both training and inference stages. Subsequently, we evaluate the \textit{accuracy} of multiple advanced methods across the DanceTrack, SportsMOT, and MOT20 benchmarks. These comparisons elucidate the distinctive characteristics of methods in different scenarios.

\begin{table}[t]
    \caption{\textbf{Efficiency comparison with TBQ methods.} `\#Parameters' refers to the total number of trainable parameters, which are expressed in millions (M). Training time corresponds to the model's training duration on DanceTrack, with the specific GPU resources clarified. Values are sourced from official publications if available. Otherwise, they are reproduced from the official implementations, which are marked by $^{\dagger}$. FPS (frames per second) of all methods is is tested \textit{consistently on one NVIDIA 4090 GPU with FP32-precision}. For MOTRv2, values of training time and FPS are shown in \textcolor{gray}{gray} to indicate they are derived from pre-computed offline detection results rather than online inference of YOLOX. Additionally, we report its practical FPS that incorporates the time of YOLOX's online inference. Similarly, we report the efficiency values of TBDQ-Net with the pretrained YOLOX for fair comparison.}
    \label{tab:efficiency}
    \centering
    \footnotesize
    \setlength{\tabcolsep}{2.6pt}
    \begin{tabular}{c|c|cc|c}
    \toprule 
    TBQ Methods & \#Parameters (M)$\downarrow$    & Training Time$\downarrow$   & GPUs      & FPS$\uparrow$   \\ \midrule
    MOTR \cite{zeng2022motr}           & 43.9                        & 60h                         & 8$\times$V100    & 14.8      \\
    MOTIP        \cite{gao2025multiple}         & 58.9                        & 36h                         & 8$\times$4090        & 14.4             \\
    CO-MOT$^{\dagger}$ \cite{yan2025comot}   & 41.7  & 125h  & 2$\times$4090   & 14.9  \\
    \multirow{2}{*}{MOTRv2$^{\dagger}$ \cite{zhang2023motrv2}}       & 41.7   & \textcolor{gray}{33h}  & 2$\times$4090        & \textcolor{gray}{20.5} \\
     & 41.7  & --  & --  & 16.0 \\  \midrule
    \rowcolor{ourscolor} \textbf{TBDQ-Net (ours)} & \textbf{8.5}  & \textbf{22h} & 2$\times$4090  & \textbf{22.0}   \\      \bottomrule
    \end{tabular}
    \vspace{-10pt}
\end{table}

\textit{Comparison in Efficiency:} As a fundamental advancement in efficiency, TBDQ-Net demonstrates dramatic superiority in efficiency over existing TBQ methods, as evidenced in Table \ref{tab:efficiency}. With only 8.5M trainable parameters, TBDQ-Net achieves approximately 80\% reduction in the size of trainable parameters compared to MOTR/MOTRv2 (41.7M+). This improvement stems from the efficient integration of pretrained detectors and a specialized MOT-tailored associator, allowing TBDQ-Net to offload computationally intensive operations from cumbersome DETR-like encoder-decoder architectures. This lightweight design also produces a significant acceleration in both training and inference. Specifically, Compared to MOTIP, TBDQ-Net reduces training resource consumption by approximately 85\% while boosting inference speed by 52\%. 
More crucially, against MOTRv2, which also uses a frozen YOLOX but retains a heavy transformer, TBDQ-Net delivers a 37.5\% faster practical inference and consumes at least 33\% fewer training resources, proving that high-performance tracking is achievable without heavy architectural redundancy. \revision{Notably, the recent DecoderTracker \cite{liao2026decodertracker} similarly strives to enhance the efficiency of query-based methods by streamlining the heavy backbone-encoder-decoder architecture. However, its advancements rely heavily on specialized backbone modifications and a complex multi-stage training pipeline. In contrast, TBDQ-Net couples a lightweight associator directly with off-the-shelf detectors, offering superior architectural flexibility and training simplicity. Overall, TBDQ-Net preserves an elegant pipeline and delivers an outstanding balance of efficiency and tracking accuracy.}

\begin{table}[t]
\centering
\footnotesize
\caption{\textbf{Comparison in DanceTrack.} The upper part of the table presents tracking-by-detection (TBD) methods, while the lower part details tracking-by-query (TBQ) methods. Methods marked with * share the same pretrained detector YOLOX. The top-2 scores are displayed in \textbf{\textcolor{blue}{blue}} and \textbf{\textcolor{green}{green}}, \textbf{\textit{aiming to represent the competitive performance of TBDQ-Net.}} -- indicates that data is not available.}
\label{results_dance}
\setlength{\tabcolsep}{3.6pt}
\begin{tabular}{c|c|ccccc}
\toprule
Methods                                                & Year            & HOTA$\uparrow$ & DetA$\uparrow$ & AssA$\uparrow$ & MOTA$\uparrow$ & IDF1$\uparrow$ \\ \hline
CenterTrack                \cite{zhou2020tracking}    & 2020                 & 41.8 & 78.1 & 22.6 & 86.8 & 35.7 \\
FairMOT                    \cite{zhang2021fairmot}    & 2021                 & 39.7 & 66.7 & 23.8 & 82.2 & 40.8 \\
ByteTrack$^{*}$   \cite{zhang2022bytetrack}           & 2022                 & 47.7 & 71.0 & 32.1 & 89.6 & 53.9 \\
FineTrack                  \cite{ren2023focus}        & 2023                 & 52.7 & 72.4 & 38.5 & 89.9 & 59.8 \\
OC-SORT$^{*}$    \cite{cao2023observation}            & 2023                 & 55.1 & 80.3 & 38.3 & \textbf{\textcolor{green}{92.0}} & 54.6 \\

ETTrack$^{*}$     \cite{han2025ettrack}  & 2024        & 56.4 & 81.7  & 39.1 & \textbf{\textcolor{blue}{92.2}}  & 57.5 \\
Hybrid-SORT$^{*}$      \cite{yang2024hybrid}      &  2024                          & 65.7 & --    & --    & 91.8 & 67.4 \\ 

MambaMOT$^{*}$    \cite{Huang2024MambaMOTSM}            &  2025             & 56.1 & 80.8  & 39.0  & 90.3 & 54.9 \\
C-TWiX$^{*}$      \cite{miah2025learning}    &  2025                        & 62.1 & 81.8 & 47.2 & 91.4 & 63.6 \\  \midrule  \midrule

MOTR                       \cite{zeng2022motr}        &  2022               & 54.2 & 73.5 & 40.2 & 79.7 & 51.5 \\
MeMOTR                     \cite{gao2023memotr}       &  2023                & 68.5 & 80.5 & 58.4 & 89.9 & 71.2 \\ 

MOTRv2$^{*}$                     \cite{zhang2023motrv2}     & 2023               & \textbf{\textcolor{green}{69.9}} & \textbf{\textcolor{blue}{83.0}} & 59.0 & 91.9 & 71.7 \\ 

CO-MOT  \cite{yan2025comot}  & 2025  & 69.4 & \textbf{\textcolor{green}{82.1}} & 58.9 & 91.2 & 71.9 \\ 

MOTIP                      \cite{gao2025multiple}     & 2025              & \textbf{\textcolor{blue}{70.0}} & 80.8 & \textbf{\textcolor{blue}{60.8}} & 91.0 & \textbf{\textcolor{blue}{75.1}} \\

\revision{DecoderTracker} \cite{liao2026decodertracker} & \revision{2026} & 54.5 & 79.0 & 35.6 & 88.1 & 54.5 \\

\rowcolor{ourscolor} $\mathrm{\textbf{TBDQ-Net}}^{*}$       & ours                    & 69.4 & 80.8 & \textbf{\textcolor{green}{59.8}} & 89.5 & \textbf{\textcolor{green}{73.4}} \\ \bottomrule
\end{tabular}
\vspace{-4pt}
\end{table}

\textit{Comparison in DanceTrack:} DanceTrack is suitable for the rigorous evaluation of target association. \textit{Despite significant efficiency gains, TBDQ-Net does not sacrifice tracking accuracy but achieves competitive performance among advanced TBQ methods}, as shown in Table \ref{results_dance}. Specifically, TBDQ-Net secures a top-3 score in the HOTA metric, highlighting its outstanding comprehensive capabilities in detection and association. When compared to MOTRv2, which adopts the same frozen detector but employs the entire MOTR model for association, our proposed associator yields higher scores in AssA (+0.8) and IDF1 (+1.7). This result highlights the efficacy of our MOT-tailored design, demonstrating that specialized interaction modules can outperform generic attention mechanisms in complex association tasks. For traditional TBD methods, their inherent weakness in association is amplified in such complex scenarios. Taking advantage of the end-to-end tracking pipeline that consists of the learnable associator and query decoding, TBDQ-Net effectively resolves this bottleneck. Consequently, TBDQ-Net achieves at least 12.0+ in AssA and 6.0+ in IDF1, with only a slight compromising in detection performance. With the same pretrained YOLOX weights, TBDQ-Net surpasses the best TBD method, Hybrid-SORT, by 3.7 points in HOTA.

\begin{table}[t]
\centering
\footnotesize
\setlength{\tabcolsep}{1.0pt}
\caption{\textbf{Comparison in SportsMOT.} The \textit{train} setting uses only the training subset, while \textit{train+val} incorporates both training and validation subsets. $\Delta H$ quantifies the HOTA improvement from the \textit{train} setting to the \textit{train+val} setting, \textbf{\textit{underscoring the model's ability to fully exploit larger data volumes}}. The best score for each metric is marked in \textbf{\textcolor{blue}{blue}}. All other notations are consistent with those presented in Table \ref{results_dance}.} 
\label{results_sports}
\begin{tabular}{c|c|ccccc|ccccc|c}
\toprule

\multirow{2}{*}{Methods} & \multirow{2}{*}{Year}  & \multicolumn{5}{c|}{\textit{train}}  & \multicolumn{5}{c|}{\textit{train+val}} & \multirow{2}{*}{$\Delta H$}   \\ \cline{3-12} 
             &        & HOTA$\uparrow$ & DetA$\uparrow$ & AssA$\uparrow$ & MOTA$\uparrow$ & IDF1$\uparrow$ & HOTA$\uparrow$ & DetA$\uparrow$ & AssA$\uparrow$ & MOTA$\uparrow$ & IDF1$\uparrow$ \\ \hline
ByteTrack$^{*}$ \cite{zhang2022bytetrack}  & 2022 & 62.8 & 77.1 & 51.2 & 94.1 & 69.8 & 64.1 & 78.5 & 52.3 & 95.9 & 71.4 & 1.3 \\
OC-SORT$^{*}$  \cite{cao2023observation}   & 2023 & 71.9 & 86.4 & 59.8 & 94.5 & 72.2 & 73.7 & 88.5 & 61.5 & 96.5 & 74.0 & 1.8 \\
MixSORT$^{*}$  \cite{cui2023sportsmot}     & 2023   & --    & --    & --    & --    & --    & 74.1 & 88.5 & 62.0 & 96.5 & 74.4 & -- \\
ETTrack$^{*}$    \cite{han2025ettrack}  & 2024  & \textbf{\textcolor{blue}{72.2}} & \textbf{\textcolor{blue}{86.9}}  & 60.1  & \textbf{\textcolor{blue}{94.9}} & 72.5  & 74.3 & \textbf{\textcolor{blue}{88.8}} & 62.1 & \textbf{\textcolor{blue}{96.8}} &  74.5 &  2.1   \\
 
MambaMOT$^{*}$    \cite{Huang2024MambaMOTSM}          & 2025   & 71.3  & 86.7  & 58.6  & \textbf{\textcolor{blue}{94.9}}  & 71.1  & -- & --   & -- & -- & -- & -- \\    \midrule \midrule
MeMOTR               \cite{gao2023memotr}   & 2023 & 70.0 & 83.1 & 59.1 & 91.5 & 71.4 & -- & -- & -- & -- & -- & -- \\
MOTIP                \cite{gao2025multiple} & 2025 & 71.9 & 83.4 & 62.0 & 92.9 & \textbf{\textcolor{blue}{75.0}} & 75.2 & 86.5 & 65.4 & 96.1 & \textbf{\textcolor{blue}{78.2}} & 3.3 \\ 
\rowcolor{ourscolor} $\mathrm{\textbf{TBDQ-Net}}^{*}$  & ours & 72.1 & 82.8 & \textbf{\textcolor{blue}{62.8}} & 89.5 & 72.9 & \textbf{\textcolor{blue}{75.5}} & 85.9 & \textbf{\textcolor{blue}{66.5}} & 92.9 & 76.1 & \textbf{\textcolor{blue}{3.4}} \\ \bottomrule

\end{tabular}
\vspace{-4pt}
\end{table}

\textit{Comparison in SportsMOT:} 
This benchmark provides two evaluation settings: the general \textit{train} setting, and the \textit{train+val} setting that further incorporates the validation set into training. As shown in Table \ref{results_sports}, TBDQ-Net achieves competitive performance in HOTA and AssA under the \textit{train} setting. When the volume of training data increases, TBDQ-Net gains the highest HOTA score. Meanwhile, TBDQ-Net demonstrates the largest HOTA improvement when transitioning from \textit{train} to \textit{train+val}, showcasing its strong data exploitation capabilities. 

Additionally, it is observed that the association performance in the above highly-dynamic datasets is exceptional. This comes from the targeted designs of our associator. Specifically, the BII-D stream actively resolves spatial redundancy by disrupting detection queries that overlap with existing tracks, preventing active tracks from being interrupted by detection intrusion. Simultaneously, the BII-T path leverages history track queries to bridge temporal gaps during occlusions, effectively ensuring the track continuity. Furthermore, the CPA module synchronizes the semantic content and positional component of queries, delivering global consistency to prevent internal semantic discrepancy. Through these synergistic mechanisms, TBDQ-Net effectively mitigates the root causes of instability, validating its efficacy in maintaining consistent object identities under complex motion dynamics.

\begin{table*}[t]
\centering
\footnotesize
\caption{\textbf{Comparison in MOT20.} Methods with $^{\dagger}$ are reproductions performed by our group. For MOTR, the number of queries was increased to 900 to better handle the crowded scenes and the training schedule is identical to ours, with all other settings adhering to the official configurations. MOTRv2 was reproduced using its official training configurations in the MOTChallenge benchmark. Scores marked in \textbf{\textcolor{blue}{blue}} are the best ones among TBQ methods, \textbf{\textit{serving to promote continued researches of query-based methods in this challenging scenario.}} All other notations are consistent with those presented in Table \ref{results_dance}.} 
\label{tab:results_mot20}
\begin{tabular}{c|c|cccccc}
\toprule
Methods                     &  Year      & HOTA$\uparrow$ & MOTA$\uparrow$ & IDF1$\uparrow$ & FP$\downarrow$ & FN$\downarrow$ & IDSW$\downarrow$ \\ \hline
FairMOT             \cite{zhang2021fairmot}         & 2021 & 54.6  & 61.8 & 67.3  & 103,440 & 88,901  & 5,243 \\
CSTrack             \cite{liang2022rethinking}      & 2022  & 54.0  & 66.6 & 68.6  & 25,404  & 144,358 & 3,196 \\
PID-MOT            \cite{lv2023one}                 & 2023   & 57.0  & 67.5 & 71.3  & 29,891  & 137,270 & 1,015 \\
MSPNet \cite{zheng2024motion}                    & 2024       & --      & 75.9 & 74.4  & 26,532  & 96,909  & 1,194  \\
GLMB   \cite{van2024visual}                        & 2024       & 54.2    & 67.7 & 67.3 & 29,597  & 134,534 &  2,911  \\
Hybrid-SORT$^{*}$      \cite{yang2024hybrid}          &  2024  & 63.9    & 76.7 & 78.4  & --     & --     & --   \\  \midrule \midrule

MOTR$^{\dagger}$    \cite{zeng2022motr}          & 2022     & 53.2 & 58.7 & 64.3  & 70,597 & 141,001 & 2,177 \\
MeMOT               \cite{cai2022memot}          & 2022     & 54.1 & 63.7 & 66.1  & 47,882 & 137,983 & 1,938 \\
MOTRv2$^{\dagger}$  \cite{zhang2023motrv2}       & 2023          & 51.7 & 51.4 & 62.8  & \textbf{\textcolor{blue}{10,111}} & 240,882 & \textbf{\textcolor{blue}{604}}   \\
\rowcolor{ourscolor} $\mathrm{\textbf{TBDQ-Net}}^{*}$    &  ours  & \textbf{\textcolor{blue}{60.2}} & \textbf{\textcolor{blue}{72.2}} & \textbf{\textcolor{blue}{72.9}} & 30,950 & \textbf{\textcolor{blue}{111,869}} & 1,200 \\ \bottomrule
\end{tabular}
\vspace{-6pt}
\end{table*}

\textit{Comparison in MOT20:} 
The severe crowd density and limited training data make MOT20 an exceptionally challenging benchmark, \textit{particularly for query-based methods as they rely on sufficient learning data to model object relationships.} Consequently, few results for TBQ methods have been publicly reported. To bridge this gap and advance the research of query-based methods under such demanding conditions, we endeavored to reproduce several state-of-the-art TBQ approaches. Our efforts confirmed that the representative methods MOTR and MOTRv2 have been successfully trained and evaluated on the official server, underscoring the inherent difficulties for TBQ methods. As shown in Table \ref{tab:results_mot20}, TBDQ-Net achieves dominant advantages across the majority of metrics, demonstrating robust association capabilities even in this challenging scenario. Although MOTRv2 obtains the best scores in FP and IDSW, it comes at the cost of a significantly inferior FN score, leading to compromised balanced tracking performance.

Generally, traditional TBD methods exhibit superior tracking performance compared to TBQ methods. This can be attributed to two primary factors:
(1) Despite the high pedestrian density, the primary motion patterns in the MOTChallenge benchmark are linear and could be effectively resolved via intricate heuristic designs. 
(2) The MOTChallenge benchmark continuously annotates and evaluates occluded, no-longer-visible pedestrians, intrinsically presenting significant hurdles to the fitting and generalization capabilities of TBQ methods.
Ultimately, the limited size of training data amplifies the two impacts, causing TBQ methods to perform inferior to TBD methods. Through this work, we hope to stimulate more attentions to these challenges, enabling query-based methods to be competent in this scenario.

\subsection{Ablation Studies}

\begin{table}[t]
\footnotesize
\caption{\textbf{Impact of each component within TBDQ-Net.} `self-attn' means the standard Transformer self-attention blocks.}
\label{ablation_each_part}
\centering
\begin{tabular}{cc|ccc}
\toprule
BII       & CPA &  HOTA & DetA & AssA \\ \hline
$\checkmark$ &   & 54.9 & 66.0 & 46.1 \\
          &$\checkmark$ & 52.9 & 65.6 & 43.1 \\
\rowcolor{ourscolor} $\checkmark$&$\checkmark$ & 58.6 & 71.4 & 48.5 \\ 
self-attn & $\checkmark$ &  57.6 & 71.1 & 47.0 \\ \bottomrule
\end{tabular}
\vspace{-4pt}
\end{table}

\textit{Analysis of TBDQ-Net Components:} We assess the individual contributions of the proposed Basic Information Interaction (BII) and Content-Position Alignment (CPA) modules through an ablation study. When the BII module is excluded, detection and track queries are directly fed into the CPA module without any prior interaction. Conversely, when the CPA module is removed, the interacted object queries are immediately decoded into tracking results, lacking the crucial alignment between content and positional components. As evident from Table \ref{ablation_each_part}, optimal performance is attained when both modules are present, demonstrating the pivotal roles of both query interaction and content-position alignment in the overall framework. Furthermore, substituting the BII module with standard Transformer self-attention blocks results in a 1.0\% drop in HOTA, suggesting the superior effectiveness of the BII module's specialized design for MOT.

\begin{table}[t]
\caption{\textbf{Impact of different configurations for noisy detection queries ($N_q$) in Equation (\ref{biidet}).}}
\label{ablation_biidet}
\centering
\footnotesize
\begin{tabular}{c|ccc}
\toprule
Configurations     & HOTA & DetA & AssA \\ \hline
track query & 57.9 & 71.1 & 47.4 \\
\rowcolor{ourscolor} hard noisy query & 58.6 & 71.4 & 48.5\\
easy noisy query & 57.6 & 71.2 & 47.0 \\
zeros       & 57.1 & 71.3 & 46.0 \\
ones        & 58.1 & 71.1 & 47.8 \\ \bottomrule
\end{tabular}
\vspace{-0.3cm}
\end{table}

\textit{Analysis of Interaction on Detection Queries (BII-D):} As elaborated in Section \ref{sec:bii}, one component of $V_1$ in Equation (\ref{biidet}) incorporates noisy queries $N_q$ to alleviate conflicts between detection queries and track queries. The impacts of various configurations for $N_q$ are evaluated in Table \ref{ablation_biidet}. The first configuration uses general track queries instead of the noisy queries, resulting in degradations in HOTA ($-0.7$) and AssA ($-1.1$). Besides, alternative configurations for $N_q$ are also explored, including easy noisy queries, an all-zero matrix, and an all-one matrix. These alternatives all yield suboptimal performance, failing to mitigate the inherent conflicts between detection queries and track queries as effectively as hard noisy queries. This is attributed to the effect of hard noisy queries, which not only disrupt potentially conflicting detection queries but also enhance the model's capacity to suppress challenging negative detections.

\begin{table}[t]
\caption{\textbf{Impact of different information sources for track query updates.} The results specifically highlight the contribution of the second terms in $K$ and $V_1$ from Equation (\ref{biitrack}).}
\label{ablation_biitrack}
\centering
\footnotesize
\begin{tabular}{c|c|ccc}
\toprule 
Track Query  & History Track Query & HOTA & DetA & AssA \\ \midrule
$\checkmark$ & $\checkmark$        & 57.3 & 71.6 & 46.2 \\ 
$\checkmark$ &                     & 57.9 & 72.1 & 46.9 \\ 
\rowcolor{ourscolor}             & $\checkmark$        & 58.6 & 71.4 & 48.5 \\ \bottomrule
\end{tabular}
\vspace{-4pt}
\end{table}

\begin{table}[t]
\caption{\textbf{Impact of different updating weights $w$ in Equation (\ref{ema}).}}
\label{ablation_hist_w}
\centering
\footnotesize
\begin{tabular}{c|ccc}
\toprule 
Updating Weight $w$  & HOTA & DetA & AssA \\ \midrule
0.9                & 58.3 & 72.0 & 47.5 \\
0.8                & 58.1 & 71.3 & 47.7 \\
\rowcolor{ourscolor} 0.7  & 58.6 & 71.4 & 48.5 \\
0.6                & 58.1 & 71.1 & 47.9 \\
\bottomrule
\end{tabular}
\vspace{-2pt}
\end{table}

\textit{Analysis of Interaction on Track Queries (BII-T):} To disclose the impact of updating track queries with different information sources, we explore various configurations for $K$ and $V_1$ in Equation (\ref{biitrack}). Detection queries, serving as the primary information source for track queries, are consistently included across all configurations. Therefore, we mainly analyze the other component of $K$ and $V_1$. As shown in Table \ref{ablation_biitrack}, suboptimal performance is achieved by either the combination of track queries and history track queries, or track queries alone. Because the former introduces redundancy in current information, while the latter suffers from a lack of crucial historical context, rendering the model less capable of robustly handling missed objects. In contrast, history track queries are able to strike an appropriate balance between current and historical information, yielding optimal performance. Furthermore, we investigate the optimal value for $w$ in Equation (\ref{ema}), which controls the update weight of new information. The results in Table \ref{ablation_hist_w} align with the intuition that current information is more relevant for detection, while historical information enhances association. Ultimately, the highest HOTA score is achieved when $w$ is set to 0.7.

\begin{table}[t]
\caption{\textbf{The performance of different values for hyperparameters $\tau_q$, $\tau_n$ and $L_{max}$.}}
\label{ablation_hyperparams}
\centering
\footnotesize
\begin{tabular}{c|ccc||c|ccc||c|ccc}
\toprule 
$\tau_q$     & HOTA & DetA & AssA   &  $\tau_n$  & HOTA & DetA & AssA   &  $L_{max}$ & HOTA & DetA & AssA \\      \midrule  \midrule
 0.2         & 57.8 & 71.8 & 46.9   &    0.4     & 57.5 & 70.9 & 47.0   &  10  & 57.3 & 72.1 & 45.9 \\
 \cellcolor{ourscolor}0.3 &  \cellcolor{ourscolor}58.6 &  \cellcolor{ourscolor}71.4 &  \cellcolor{ourscolor}48.5   &  \cellcolor{ourscolor}0.5   &  \cellcolor{ourscolor}58.6 &  \cellcolor{ourscolor}71.4 &  \cellcolor{ourscolor}48.5   &  15  & 57.7 & 71.7 & 46.8 \\
 0.4         & 58.0 & 71.3 & 47.5   &    0.6     & 58.2 & 70.7 & 48.2   &  \cellcolor{ourscolor}20  &  \cellcolor{ourscolor}58.6 &  \cellcolor{ourscolor}71.4 &  \cellcolor{ourscolor}48.5 \\
 0.5         & 57.4 & 71.9 & 46.2   &    0.7     & 55.6 & 67.6 & 46.1   &  25  & 57.8 & 71.2 & 47.3  \\      
\bottomrule
\end{tabular}
\end{table}

\textit{Analysis of Other Hyperparameters:} The impacts of different values for hyperparameters $\tau_q$ (Section \ref{sec:bii}), $\tau_n$ and $L_{max}$ (Section \ref{sec:traininginference}) in DanceTrack are evaluated in Table \ref{ablation_hyperparams}. Although performance fluctuations are observed with different parameter settings, these variations are generally within an acceptable range. For SportsMOT and MOT20, empirical studies show that dataset-specific tuning of these hyperparameters could yield higher performance. However, we maintain consistency with the optimal configurations in DanceTrack for conciseness.

\begin{table}[t]
\caption{\textbf{The performance of different augmentation strategies for clip length, random insert ($p_i$) and random dropout ($p_d$) in MOT20.}}
\label{ablation_aug_clip_length}
\setlength{\belowcaptionskip}{-0.2cm}
\centering
\footnotesize
\begin{tabular}{c|ccc||c|ccc}            \toprule
Clip Length & HOTA   & DetA & AssA   & $(p_i, p_d)$ & HOTA   & DetA & AssA    \\  \midrule
 6          & 61.4   & 63.3 & 59.8   &   (0.0, 0.0) &  61.9  & 55.9 & 68.5  \\
 7          & 62.9   & 64.5 & 61.5   &    \cellcolor{ourscolor}(0.1, 0.1) &   \cellcolor{ourscolor}72.3  &  \cellcolor{ourscolor}72.5 &  \cellcolor{ourscolor}72.2  \\
 8          & 70.1   & 70.3 & 70.0   &   (0.1, 0.2) &  63.4  & 65.0 & 62.0  \\
 \cellcolor{ourscolor}9          &  \cellcolor{ourscolor}71.6   &  \cellcolor{ourscolor}74.4 &  \cellcolor{ourscolor}69.2   &   (0.2, 0.1) &  71.6  & 74.4 & 69.2  \\
 10         & 64.7   & 65.5 & 64.1   &   (0.2, 0.2) &  60.8  & 65.0 & 57.1  \\  \bottomrule
\end{tabular}
\vspace{-4pt}
\end{table}

\textit{Ablation Studies in MOT20:} Table \ref{ablation_aug_clip_length} quantitatively validates the necessity of our specific strategies for the scarce data of the MOT20 benchmark. As shown in the left columns, extending the clip length yields substantial gains. For instance, increasing it from 6 to 9 boosts HOTA by +10.2 points, confirming that longer temporal context contributes to the optimization when training data is limited. The right columns demonstrate that introducing random insertion ($p_i$) and dropout ($p_d$) further enhances robustness. The optimal setting ($p_i=0.1$, $p_d=0.1$) improves HOTA by +10.4 points compared to the baseline without augmentation. These operations effectively simulate challenging cases such as false alarms and occlusions, preventing the model from overfitting to the simple patterns in the small training set.

\begin{table}[t]
\footnotesize
\caption{\textbf{Integration with different detectors.}}
\label{ablation_detector}
\centering
\begin{tabular}{c|c|cccc}
\toprule 
Detector  & Architecture & HOTA & DetA & AssA & mAP \\ \midrule
DAB-DETR \cite{liu2022dabdetr} & Transformer  & 58.6 & 71.4	& 48.5	& 68.9 \\
RT-DETR \cite{zhao2024detrs} & Transformer & 60.6 & 73.5 & 50.2 & 79.4 \\
YOLOX \cite{ge2021yolox} & CNN & 63.2 &	76.8 &	52.2 &	81.2 \\

\bottomrule
\end{tabular}
\vspace{-2pt}
\end{table}

\textit{Compatibility with Diverse Detectors:} A key advantage of our decoupled design is that the associator is architecture-agnostic. It operates on generic detection priors and query interactions, making it compatible with various detection models. To verify this versatility, we integrate our associator with two distinct Transformer-based detectors, DAB-DETR \cite{liu2022dabdetr} and RT-DETR \cite{zhao2024detrs}, alongside the CNN-based YOLOX \cite{ge2021yolox}. As shown in Table \ref{ablation_detector}, the framework adapts seamlessly to different architectures. The tracking performance (HOTA) generally aligns with the detection quality (mAP), indicating that our associator is robust to detection architectures and benefits from stronger detection models.

\begin{figure*}[htbp] 
    \centering
    \begin{minipage}{\linewidth}
        \centering
        \includegraphics[width=0.9\linewidth]{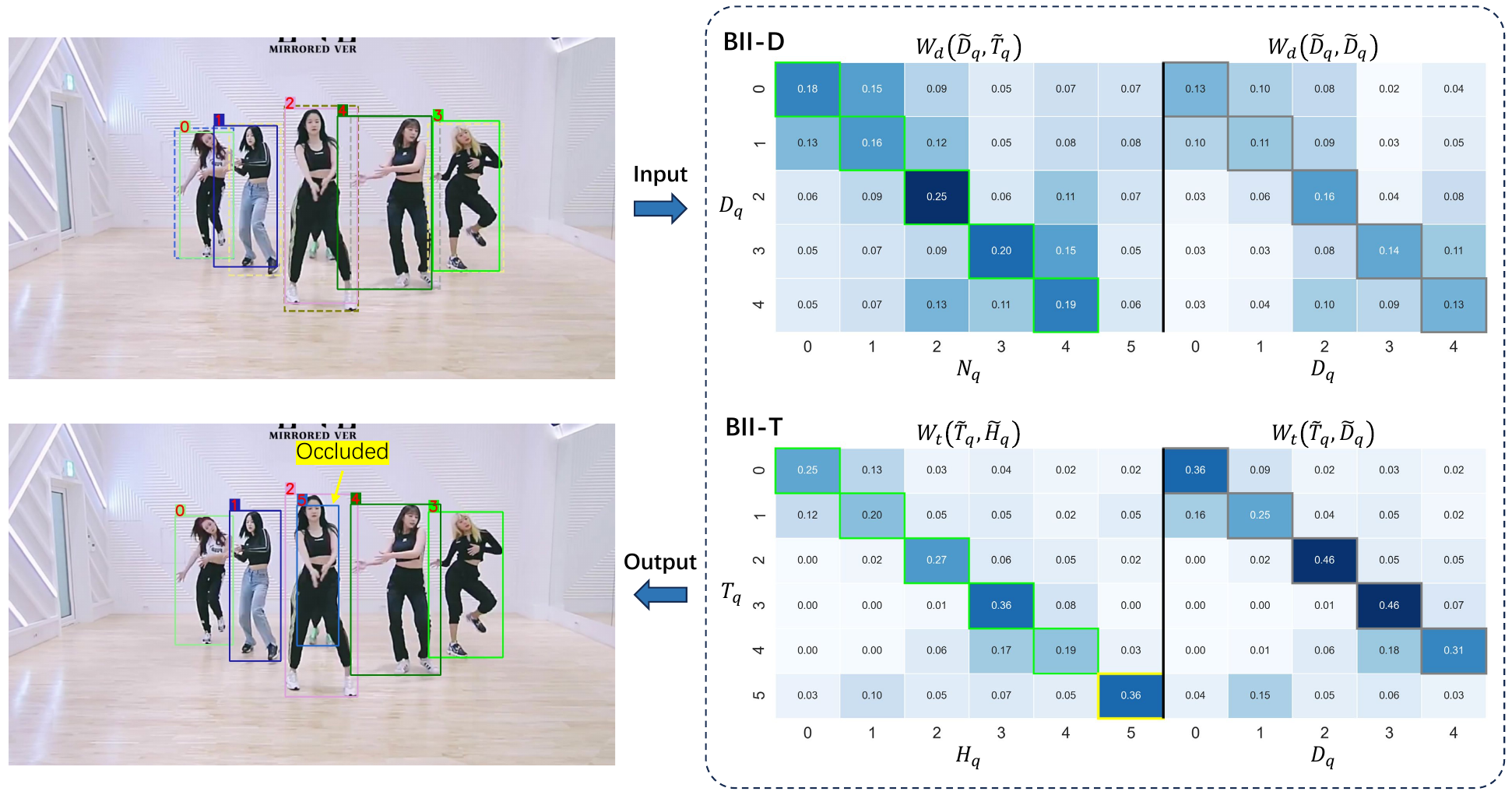}
        \\[-0.8em] 
        \centerline{\footnotesize (a) Case 1: Conflict suppression and occlusion handling.}
    \end{minipage}
    
    \vspace{0.8em} 
    
    \begin{minipage}{\linewidth}
        \centering
        \includegraphics[width=0.9\linewidth]{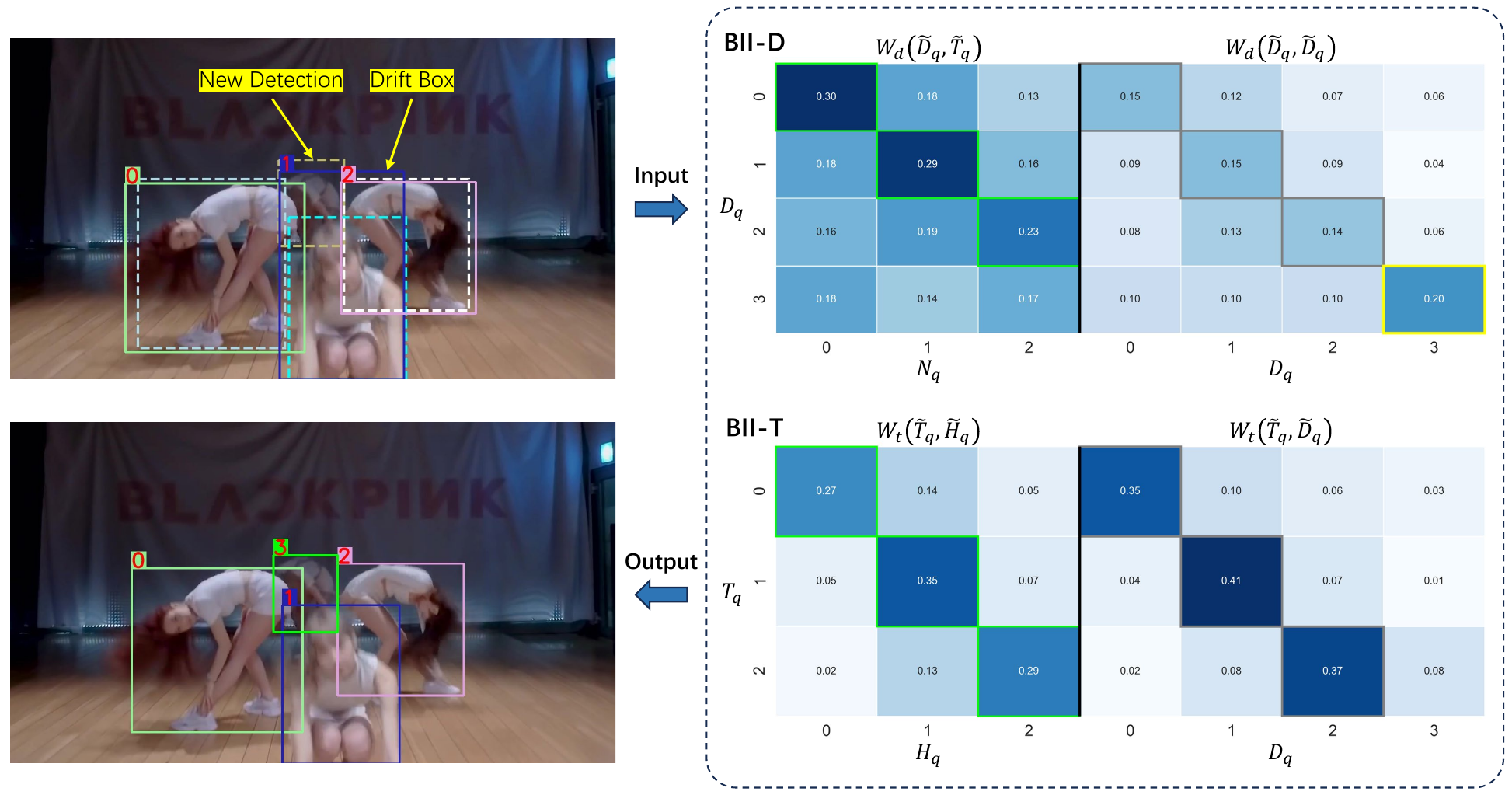}
        \\[-0.8em]
        \centerline{\footnotesize (b) Case 2: Drift correction and new track initiation.}
    \end{minipage}
    \\[-0.8em]
    \caption{
    \textbf{Visualization of attention weights in the BII dual-stream.} In each case, the left part presents a tracking scenario. The dashed and solid bounding boxes represent current-frame detections and previous tracking results, respectively. Yellow highlights indicate noteworthy individuals. The right part illustrates the attention matrices. $W_d$ and $W_t$ visualize the attention weights between Query and Key components as defined in Equation \ref{biidet} and \ref{biitrack}. The matrix indices $(0, 1, ...)$ correspond to the object IDs shown in the left images.
    }
    \label{fig:bii_cases}
\end{figure*}

\subsection{Visualizations}
\textit{Visualization of the BII Mechanism:} We clarify the working mechanism of the BII module by visualizing the attention weights in two representative scenarios, as shown in Figure \ref{fig:bii_cases}. In Case 1, severe spatial conflict arises when a detection box overlaps with a tracked object. The attention matrix $W_d(\widetilde{D_q}, \widetilde{T_q})$ exhibits prominent weights along the diagonal, demonstrating that noisy queries $N_q$ effectively disrupt corresponding detection queries by exerting more impacts, thereby avoiding the competitions. Meanwhile, current detection information is transferred to track queries via $W_t(\widetilde{T_q}, \widetilde{D_q})$ for real-time state updates, while $W_t(\widetilde{T_q}, \widetilde{H_q})$ supplies temporal stability via history track queries $H_q$. Notably, dancer \#5 is missing from the current detections due to occlusion. However, by leveraging $H_q$, the framework successfully maintains track continuity despite the absence of localization priors in the current frame.

A similar mechanism is observed in Case 2, where noisy queries suppress redundant detection queries and history track queries stabilize the tracking process. It is noted that track \#1 initially suffers from spatial drift caused by object crowding and large displacements. By promptly integrating current detection observations, its position is successfully corrected, ensuring high tracking precision. Furthermore, a new detection is correctly identified and initialized as a new identity, ID \#3. It is not suppressed by noisy queries since there is no spatial conflict with existing tracks.

Overall, the visualizations align well with the original motivation of the BII module. Although the attention distribution may show minor fluctuations, such as high activations outside the highlighted regions, these variations do not compromise the final outcome, which benefits from the inherent robustness of the neural network system. In particular, these optimized attention weights share conceptual similarities with the affinity matrices used in TBD methods. \textit{However, a fundamental distinction exists in that TBDQ-Net does not use these weights for explicit, hard assignment. Instead, it harnesses them to facilitate learnable, soft interactions among object queries.} These refined queries are subsequently decoded into tracking results, enabling the model to gain a more holistic understanding of the environment and capture robust tracking cues beyond simple heuristic matching.

\begin{figure}[t]
    \centering
    \footnotesize
    \includegraphics[width=1.0\linewidth]{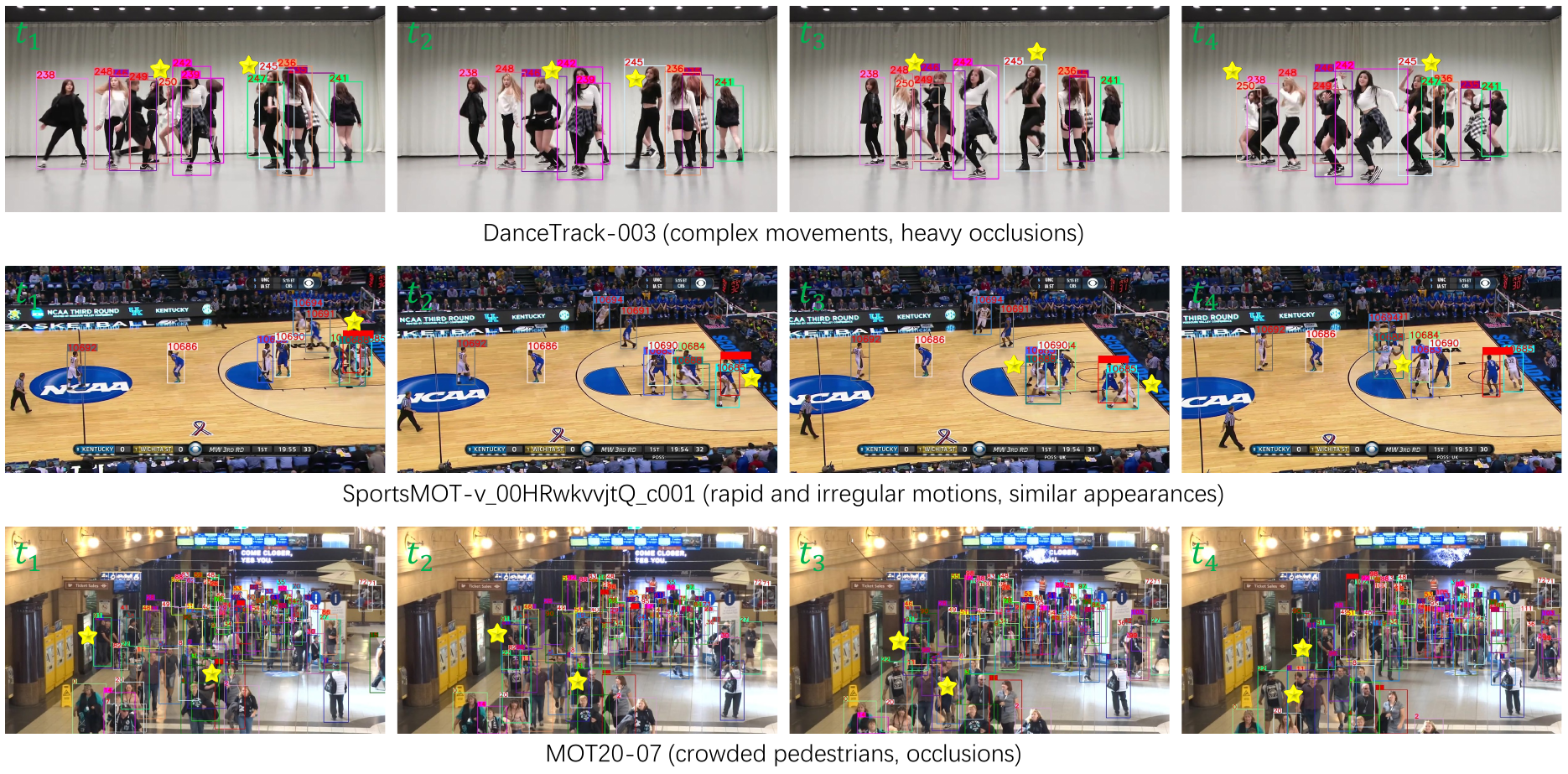}
    \caption{\textbf{Visualization of tracking results in diverse challenging scenarios.} Yellow stars signify the noteworthy individuals in each frame. Object IDs are discernible upon magnified viewing.}
    \label{fig:vis_all}
\end{figure}

\textit{Visualization of Tracking Results:} The tracking results of TBDQ-Net are visualized in Figure \ref{fig:vis_all}. In DanceTrack, clustering and occlusion occur among dancers \#246, \#249, and \#250; however, they are successfully tracked upon reappearance. A similar phenomenon is observed for dancer \#247, who is heavily occluded by dancer \#245, but is promptly re-identified when reappearing. In SportsMOT, TBDQ-Net effectively tracks players with rapid, irregular motions and visually similar appearances. In MOT20, TBDQ-Net consistently tracks pedestrian \#6, despite spatial overlaps with surrounding individuals in extremely crowded environments. Similarly, it accurately differentiates between pedestrians \#81 and \#91. These results highlight the robust modeling capabilities of TBDQ-Net in complex scenarios.

\begin{figure*}[]
    \centering
    \footnotesize
    \setlength{\abovecaptionskip}{-0.3pt}
    \setlength{\belowcaptionskip}{-6pt}
    \includegraphics[width=1.0\linewidth]{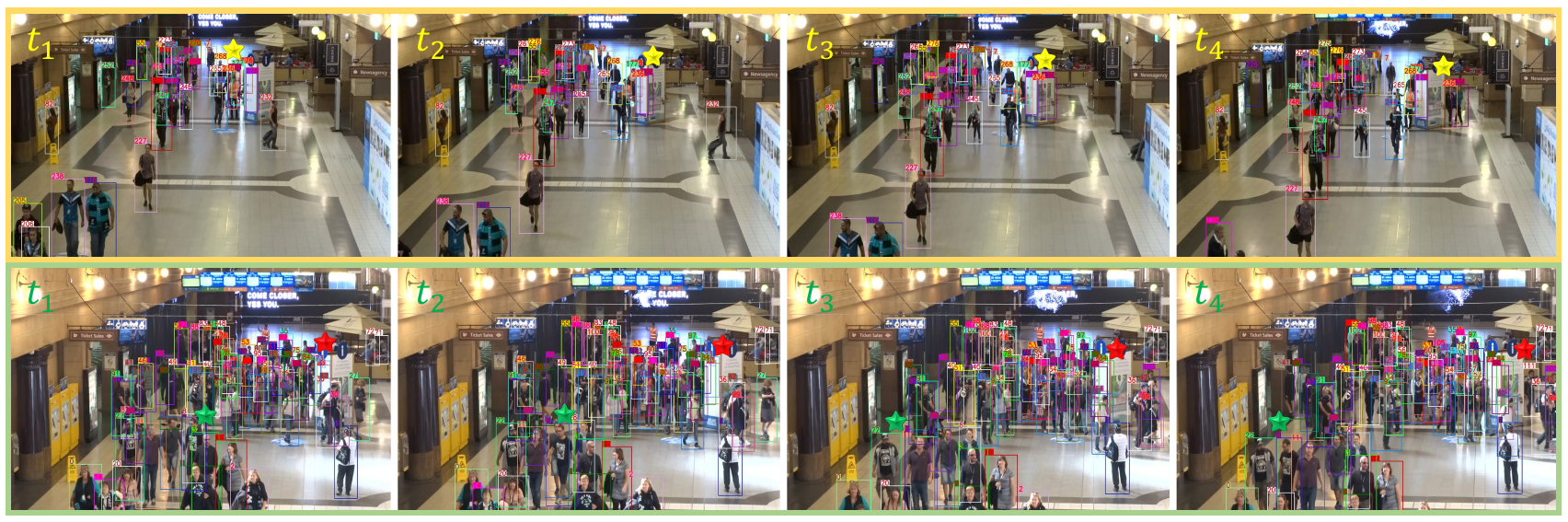}
    \caption{Visualization of typical failure cases in MOT20. Top row: Ground truth annotations from the training set. Pedestrians \#235 and \#236 (yellow stars) are fully occluded during $t_2 \to t_3$ but are annotated via linear interpolation between their visible states at $t_1$ and $t_4$, contrasting with the stationary pedestrian \#178. This annotation style implies an offline tracking paradigm. Bottom row: Tracking results from the test set. Pedestrian \#84 (red star), undergoing similar occlusion, suffers an ID switch upon reappearance at $t_4$ (\#111). In contrast, pedestrian \#6 (green star), despite severe overlap ($t_2$) and occlusion ($t_3$), is successfully tracked, demonstrating the model's robustness under standard online conditions.
    }
    \label{fig:case3}
\end{figure*}

\subsection{Discussion and Limitations}

\textbf{Dependency on Detector Quality:} TBDQ-Net demonstrates strong association capabilities with remarkable efficiency. However, given that we freeze the detector, its detection quality may still be less competitive. Therefore, improving detection quality remains a valuable direction.

\textbf{Scope of Generalization:} It is worth noting that the demonstrated advantages of TBDQ-Net are most pronounced on human-centric and high-density datasets. Nonetheless, performance may vary on differently structured benchmarks, which potentially stem from differences in camera dynamics (e.g., dominant ego-motion), annotation rules (e.g., occlusion handling), and data distribution. Consequently, applying TBDQ-Net to these distinct domains might require domain-specific adaptations to the query interaction modules to maintain robustness.

\textbf{Typical Failure Cases:} Despite robust performance, TBDQ-Net faces challenges in scenarios involving fully invisible targets under offline annotation rules. As illustrated in Figure \ref{fig:case3} (top), in the MOT20 ground truth, pedestrians \#235 and \#236 are completely occluded by a billboard but are annotated via linear interpolation. This assumes a consistent motion model that is indistinguishable from stationary objects (e.g., pedestrian \#178) without future information, inherently favoring offline paradigms. In contrast, TBDQ-Net operates in an online manner. As shown in Figure \ref{fig:case3} (bottom), pedestrian \#84, undergoing similar occlusion, suffers an ID switch upon reappearance. However, under standard challenges such as the severe visual overlap and occlusion experienced by pedestrian \#6, our model successfully maintains identity consistency. This comparison highlights that handling the specific invisible-but-annotated cases in benchmarks like MOT20 requires incorporating long-term motion prediction, specific re-identification feature bank, or offline refinement strategies, which is a promising direction for future improvement.

\textbf{Data Dependency:} As a query-based framework with a learnable associator, TBDQ-Net tends to benefit more from data-rich environments. This characteristic makes it well-suited for modern, large-scale applications, such as DanceTrack and SportsMOT. But it also introduces a dependency on sufficient training data. For instance, under settings where training data is limited (e.g., MOT20), our method currently trails heuristic-based approaches that are more data-efficient. Future work could explore few-shot learning strategies to mitigate this.

\section{Conclusion}

In this work, we present TBDQ-Net, a unified multi-object tracking framework designed to foster the synergy between the decoupled tracking-by-detection (TBD) and the end-to-end tracking-by-query (TBQ) paradigms. By structurally integrating a frozen detector with a lightweight associator, we demonstrate that specialized association mechanisms can achieve robust tracking performance within this streamlined query-based architecture. Specifically, the task-tailored Basic Information Interaction (BII) and Content-Position Alignment (CPA) modules ensure effective tracking while eliminating the substantial overhead of model redundancy and joint training.

Functionally, TBDQ-Net validates that effective association in challenging scenarios, such as the non-linear motion in DanceTrack and SportsMOT or the ultra-dense crowds in MOT20, can be achieved through specialized lightweight designs. This significantly lowers the engineering barrier for end-to-end tracking and enables the flexible integration of diverse high-performance detectors, thus offering strong practical utility.

Regarding limitations, the frozen-detector strategy, while ensuring structural efficiency, restricts further refinement of detection features, making TBDQ-Net influenced by pre-trained detection performance. In addition, as a query-based framework, its optimal performance exhibits a dependency on data-rich environments; in scenarios with extremely limited training data or annotation protocols requiring persistent tracking of completely invisible targets, heuristic-based methods with scene-specific optimizations currently remain advantageous.

Looking forward, these limitations outline clear directions for future work, including lightweight adapter mechanisms to alleviate detection bottlenecks, more data-efficient query interaction strategies or external knowledge integration (e.g., Vision-Language Models), and extensions of the TBDQ paradigm to multi-modal tracking by incorporating LiDAR or depth cues. We hope TBDQ-Net serves not only as a strong baseline but also as a useful reference for advancing efficient and unified tracking architectures.

\section*{Acknowledgements}
We are grateful for the support of the National Natural Science Foundation of China (No. 62271143), the Frontier Technologies R\&D Program of Jiangsu (No.BF2024060), and the Big Data Computing Center of Southeast University. \revision{We also thank the anonymous reviewers for their valuable comments and suggestions.}

\section*{Declaration of Generative AI}
During the preparation of this work, the authors used ChatGPT to refine the academic phrasing, improve the overall grammatical structure of the manuscript, and ensure correct spelling and formatting. After using this tool, the authors thoroughly reviewed and edited the content as needed and take full responsibility for the ultimate content of the publication.

\bibliographystyle{elsarticle-num}
\bibliography{ref.bib}

\end{document}